\documentclass{article}

\PassOptionsToPackage{numbers, compress}{natbib}

\usepackage[preprint]{neurips_2021}
\usepackage{dsfont}

\usepackage[utf8]{inputenc} 
\usepackage[T1]{fontenc}    
\usepackage{hyperref}       
\usepackage{url}            
\usepackage{booktabs}       
\usepackage{amsfonts}       
\usepackage{nicefrac}       
\usepackage{microtype}      
\usepackage{xcolor}         
\usepackage{graphicx}
\usepackage{amsmath,bm}
\usepackage{algorithmicx}
\usepackage[ruled]{algorithm}
\usepackage[noend]{algpseudocode}
\usepackage{wrapfig}
\algrenewcommand\algorithmicrequire{\textbf{Input:}}
\algrenewcommand\algorithmicensure{\textbf{Output:}}

\usepackage{mathtools}
\newcommand{\N}{\mathcal{N}}

\newcommand*\samethanks[1][\value{footnote}]{\footnotemark[#1]}

\DeclarePairedDelimiterX{\infdivx}[2]{(}{)}{%
  #1\;\delimsize\|\;#2%
}

\DeclareMathOperator{\EX}{\mathbb{E}}

\title{Input Dependent Sparse Gaussian Processes}

\author{Bahram Jafrasteh\thanks{Equal contribution} \\ 
		Computer Science Department\\  Universidad Aut\'onoma de Madrid \\ bahram.jafrasteh@uam.es
		\And Carlos Villacampa-Calvo\samethanks[1] \\
		Computer Science Department\\  Universidad Aut\'onoma de Madrid \\ carlos.villacampa@uam.es
		\And Daniel Hern\'andez-Lobato \\
		Computer Science Department\\  Universidad Aut\'onoma de Madrid \\ daniel.hernandez@uam.es}

\begin{document}

\maketitle

\begin{abstract}
Gaussian Processes (GPs) are Bayesian models that provide uncertainty 
estimates associated to the predictions made. They are also very flexible 
due to their non-parametric nature. Nevertheless, GPs suffer from poor 
scalability as the number of training instances $N$ increases. 
More precisely, they have a cubic cost with respect to $N$. To overcome 
this problem, sparse GP approximations are often used, where a set of 
$M \ll N$ inducing points is introduced during training. The location of 
the inducing points is learned by considering them as parameters of an 
approximate posterior distribution $q$. Sparse GPs, combined with variational 
inference for inferring $q$, reduce the training cost of GPs to $\mathcal{O}(M^3)$. 
Critically, the inducing points determine the flexibility of the model and 
they are often located in regions of the input space where the latent 
function changes. A limitation is, however, that for some learning tasks 
a large number of inducing points may be required to obtain a good 
prediction performance. To address this limitation, we propose here 
to amortize the computation of the inducing points locations, as well 
as the parameters of the variational posterior approximation $q$. For this, 
we use a neural network that receives the observed data as an input and 
outputs the inducing points locations and the parameters of $q$. We evaluate 
our method in several experiments, showing that it performs similar or better 
than other state-of-the-art sparse variational GP approaches. However, 
with our method the number of inducing points is reduced drastically due 
to their dependency on the input data. This makes our method scale 
to larger datasets and have faster training and prediction times. 
\end{abstract}

\section{Introduction}
\label{sect:intro}

Gaussian Processes (GPs) are non-parametric models that can be 
used to address machine learning problems, including regression and 
classification \citep{rasmussen2005book}. GPs become more expressive as 
the number of training instances $N$ grows and, since they are Bayesian models, 
they provide a predictive distribution that estimates the uncertainty associated 
to the predictions made. This uncertainty estimation or ability to know that is not 
known becomes critical in many practical applications \cite{gal2016uncertainty}. 
Nevertheless, GPs suffer from poor scalability as their training cost is $\mathcal{O}(N^3)$ due to 
the need of computing the inverse of a covariance matrix of size $N\times N$. 
Another limitation is that approximate inference is required with non-Gaussian likelihoods \cite{rasmussen2005book}.

To improve the cost of GPs, sparse approximations can be used
\cite{rasmussen2005book}. The most popular ones introduce
a set of $M \ll N$ inducing points \cite{Snelson2006,titsias_variational_2009}.
The inducing points and their associated posterior values completely specify the 
posterior process at test points. In the method described in \cite{Snelson2006}, the 
computational gain is obtained by assuming independence among the process values at the
training points given the inducing points and their values. This approach can also be seen 
as using an approximate GP prior \citep{quinonero2005unifying}.
By contrast, in the method in \cite{titsias_variational_2009} the computational gain is 
obtained by combining variational inference (VI) with a posterior approximation $q$ that has a 
fixed part and a tunable part. In both methods the computational cost is reduced 
to $\mathcal{O}(NM^2)$ and the inducing points, considered as model's hyper-parameters,
are learned by maximizing an estimate of the marginal likelihood. 

Importantly, the VI approach in \cite{titsias_variational_2009} maximizes a lower 
bound on the log-marginal likelihood as an indirect way of minimizing the KL-divergence 
between an approximate posterior distribution for the process values associated to the 
inducing points and the corresponding exact posterior. The advantage 
of this approach is that the objective is expressed as a sum over the training instances, 
allowing for mini-batch training and stochastic optimization techniques to be applied on 
the objective \cite{hensman2015}. This reduces the total training cost to $\mathcal{O}(M^3)$, 
making GPs scalable to very large datasets.

When using sparse approximations, however, one often observes in practice that
after the optimization process the inducing points are located in those regions of the 
input space in which the latent function changes \cite{Snelson2006,titsias_variational_2009,HensmanMG15,bauer2016}.
Therefore, the expressive power of the model critically depends on the number of inducing
points $M$ and on its correct placement across the input space. For example, some
problems may require a large number of inducing points, in the order of several hundreds, 
to get good prediction results \cite{hensman2015,shi2020sparse,tran2020sparse}. 
This makes difficult using sparse GPs based on inducing points in those problems.

In the literature there have been some attempts to improve the computational cost of
sparse approximations. These approaches follow different strategies, including 
using different sets of inducing points for the computation of the posterior mean
and variance \cite{cheng2017}. Other approaches use an orthogonal decomposition
of the GP that allows to introduce an extra set of inducing points with less cost 
\cite{shi2020sparse}. Finally, other methods consider a large set of inducing points, 
but restrict the computations for a particular data point to the nearest neighbors
to that point from the set of inducing points \cite{tran2020sparse}.

In this work we are inspired by the approach of \cite{tran2020sparse} and 
propose a novel method to improve the computational cost of sparse GPs. 
Our method also tries to produce a set of inducing points (and associated variational approximation $q$) 
that are specific of each different input data point, as in \cite{tran2020sparse}.  For that, we note that
some works in the literature have observed that one can learn the mappings from inputs to proposal 
distributions instead of directly optimizing their parameters \citep{KingmaW13, shu2018amortized}. This 
approach, known as amortized variational inference, is a key contribution of variational auto-encoders 
(VAE) \citep{KingmaW13}, and has also been explored in the context of GP to solve other 
types of problems such as multi-class classification with input noise \citep{villacampa21}.
Amortized inference has also been empirically shown to lead to useful regularization properties
that improve the generalization performance \citep{shu2018amortized}.

Specifically, here we propose to combine sparse GP with a neural network architecture to compute the inducing 
points locations associated to each input point. Moreover, we also employ a neural network to carry out amortized VI 
to compute the parameters of the approximate variational distribution $q$ modeling the posterior distribution
associated to the values of the outputted inducing points. Critically, this approach allows to reduce the number of inducing points 
drastically without loosing expressive power, as now we have different sets of inducing points associated to each input location. 
The inducing points are simply given by a mapping from the inputs provided by a neural network. We show 
on several experiments that the proposed method is able to perform similar or better than standard sparse GPs 
and competitive methods for improving the cost of sparse GPs \citep{tran2020sparse,shi2020sparse}. 
However, the training and prediction times of our method are much better.

\section{Gaussian processes}
\label{sect:gp}

A Gaussian Process (GP) is a stochastic process for which any finite set of variables has 
a Gaussian distribution \cite{rasmussen2005book}. GPs are non-linear and non-parametric approaches 
for regression and classification. In a learning task, we use a GP as a prior over a latent function.  
Then, Bayes' rule is used to get a posterior for that function given the observed data. Consider a dataset
$\mathcal{D} =\left\{ \left(\mathbf{x}_i, y_i \right) \right\}_{i=1}^N$, where each 
scalar $y_{i}$ is assumed to be obtained by adding an independent and identically distributed Gaussian noise with 
a zero mean and variance $\sigma^2$ to a function $f(\cdot)$ evaluated 
on $\mathbf{x}_i$, \emph{i.e.}, $y_{i} = f(\mathbf{x_{i}}) + \epsilon_i$, where 
$\epsilon_i \sim \N(0, \sigma^{2})$. We specify a prior distribution for $f$ 
in the form of a GP, which is described by a mean function $m(\mathbf{x})$ (often set to zero) and covariance function 
$k(\mathbf{x}, \mathbf{x}')$ such that $k(\mathbf{x}, \mathbf{x}')=\mathds{E}[f(\mathbf{x})f(\mathbf{x}')]$. 
Covariance functions typically have some adjustable parameters $\theta$.
The GP formulation allows to compute a posterior or predictive distribution for the potential values of $f$ 
given $\mathcal{D}$. More precisely, the the prediction at a new test point $\mathbf{x}^\star$ is 
Gaussian with mean and variance given by
\begin{align}
\label{Eq.GPmean}
\mu(\mathbf{x}^\star) &= \mathbf{k}(\mathbf{x}^\star)^\text{T} (\mathbf{K}+\mathbf{\sigma^{2} \mathbf{I}})^{-1} \mathbf{y}\,,
\\
\sigma^{2}(\mathbf{x}^\star) & = 
	k(\mathbf{x}^\star,\mathbf{x}^\star) - \mathbf{k}(\mathbf{x}^\star)^\text{T} (\mathbf{K}+\mathbf{\sigma^{2} \mathbf{I}})^{-1} 
	\mathbf{k}(\mathbf{x}^\star)\,,
\label{Eq.GPvar}
\end{align}
where $\mu(\mathbf{x^\star})$ and $\sigma^{2}(\mathbf{x}^\star)$ are the prediction mean and variance, 
respectively. $\mathbf{k}(\mathbf{x}^\star)$ is a vector with the covariances between $f(\mathbf{x}^\star)$ 
and each $f(\mathbf{x}_i)$. Similarly, $\mathbf{K}$ has the covariances between $f(\mathbf{x}_i)$ and $f(\mathbf{x}_j)$
for $i,j=1,\ldots, N$. Finally, $\mathbf{I}$ stands for the identity matrix.
A popular covariance function $k(\cdot,\cdot)$ is the squared exponential covariance function, which assumes 
that $f$ is smooth \cite{rasmussen2005book}. Its parameters, $\theta$, and $\sigma^2$ can simply be found via 
type-II maximum likelihood estimation  by maximizing $p(\mathbf{y})$ \cite{rasmussen2005book}.
Note, however, that the computational complexity of this approach is $O(N^{3})$ since it 
needs the inversion of $\mathbf{K}$, a $N \times N$ matrix. This makes GPs unsuitable 
for large data sets. 

\subsection{Sparse variational Gaussian processes}
\label{sect:sparse_gp}

Sparse Gaussian process improve the computational cost of GPs. The most popular approaches 
introduce, in the same input space as the original data, a new set of $M \ll N$ points , called 
the inducing points, denoted by $\mathbf{Z}=(\mathbf{z}_1,\ldots,\mathbf{z}_M)^\text{T}$ \citep{Snelson2006,titsias_variational_2009}. 
Let the corresponding latent functions be $\mathbf{u}=(f(\mathbf{z}_1),\ldots, f(\mathbf{z}_M))^\text{T}$. 
The inducing points are not restricted to be part of the observed data and 
their location can be learned during training.  A GP prior is placed on $\mathbf{u}$. 
Namely, $p(\mathbf{u}) \sim \mathcal{N}(\mathbf{0}, \mathbf{K}_\mathbf{Z})$, 
where $\mathbf{K}_\mathbf{Z}$ is a matrix with the covariances associated to each pair of points from $\mathbf{Z}$. 
The main idea of sparse approximations is that 
the posterior for $f$ can be approximated in terms of the posterior for $\mathbf{u}$.

In this work we focus on a widely used variational inference (VI) approach to 
approximate the posterior for $f$ \citep{titsias_variational_2009}. 
Let $\mathbf{f}=(f(\mathbf{x}_1),\ldots,f(\mathbf{x}_N))^\text{T}$.
In VI, the goal is to find an approximate posterior 
for $\mathbf{f}$ and $\mathbf{u}$, $q(\mathbf{f}, \mathbf{u})$, 
that resembles as much as possible the true posterior $p(\mathbf{f},\mathbf{u}|\mathbf{y})$. 
Critically, $q$ is constrained to be $q(\mathbf{f}, \mathbf{u})=p(\mathbf{f}|\mathbf{u})q(\mathbf{u})$,
with $p(\mathbf{f}|\mathbf{u})$ fixed and $q(\mathbf{u})$ a tunable multi-variate Gaussian.
To find $q(\mathbf{u})$ a lower bound of the marginal likelihood is maximized. This 
bound is obtained through Jensen's inequality, leading to the following expression,
after some simplifications:
\begin{align}
\mathcal{L} = \sum_{i=1}^{N}\EX_{q(\mathbf{f})}[\log p(y_i|f_i)] - 
	\text{KL}[q(\mathbf{u})|p(\mathbf{u})]\,,
	\label{eq:elbo_vi}
\end{align}
where $p(\mathbf{y}|\mathbf{f})$ is the model's likelihood and $\text{KL}[\cdot|\cdot]$ is the Kullback-Leibler
divergence between probability distributions. In \cite{titsias_variational_2009}, they do optimize $q(\mathbf{u})$ 
in closed-form. The resulting expression is then maximized to estimate $\mathbf{Z}$, $\theta$ and $\sigma^2$.
This leads to a complexity of $\mathcal{O}(NM^2)$. However, if 
the variational posterior $q(\mathbf{u})$ is optimized alongside with $\mathbf{Z}$, $\theta$ and $\sigma^2$,
as proposed in \cite{hensman2013}, the ELBO can be expressed as a sum over training instances,
which allows for mini-batch training and stochastic optimization techniques. 
Using stochastic variational inference (SVI) reduces the training cost to $\mathcal{O}(M^3)$ \cite{hensman2013}.
Importantly, the first term in (\ref{eq:elbo_vi}) is an expectation that has closed-form solution 
in the case of Gaussian likelihoods. It needs to be approximated for other cases, \emph{e.g.}, binary classification, 
either by quadrature or MCMC methods \cite{hensman2015}. The second term is the KL-divergence 
between the variational posterior and the prior, which can be computed analytically since they are both Gaussian.

After optimizing (\ref{eq:elbo_vi}), one often observes that
the inducing points are located in those regions of the input space 
in which the latent function changes \cite{titsias_variational_2009,HensmanMG15,bauer2016}.
In consequence, the expressive power of the sparse GP critically depends on the number of inducing
points $M$ and on its correct placement. In some learning problems, however, 
a large number of inducing points, in the order of several hundred, is required 
to get good prediction results \cite{hensman2015,shi2020sparse,tran2020sparse}. 
This makes difficult and expensive using sparse GPs in those problems.
In the next section we describe how to reduce the cost of this method.

\section{Input dependent sparse GPs}
\label{sect:input_dep_gp}

We develop a novel formulation of sparse GPs, which for every 
given input computes the corresponding inducing points to be used
for prediction, and also the associated parameters of the approximate
distribution $q$. To achieve this goal we consider a meta-point 
$\tilde{\mathbf{x}}$ that is used to determine the inducing points 
$\mathbf{Z}$ and the corresponding $\mathbf{u}$. 
Namely, now $\mathbf{u}$ depends on $\tilde{\mathbf{x}}$, \emph{i.e.}, $\mathbf{u} \sim p(\mathbf{u}|\tilde{\mathbf{x}})$.
In particular, we set $p(\mathbf{u}| \tilde{\mathbf{x}}) = \mathcal{N}(\mathbf{0}, 
\mathbf{K}_{\mathbf{Z}(\tilde{\mathbf{x}})})$ where the inducing points 
$\mathbf{Z}$ depend non-linearly, \emph{e.g.}, via a deep neural network, on $\mathbf{\tilde{x}}$.
The joint distribution of $\mathbf{u}$ and $\tilde{\mathbf{x}}$ is then given by
$p(\mathbf{u},\tilde{\mathbf{x}})=p(\mathbf{u}|\tilde{\mathbf{x}})p(\tilde{\mathbf{x}})$
for some prior distribution $p(\tilde{\mathbf{x}})$ over $\tilde{\mathbf{x}}$.
Following \citep{tran2020sparse}, we can consider an implicit distribution 
$p(\tilde{\mathbf{x}})$. That is, its analytical form is 
unknown, but we can draw samples from it. Later on, we will specify $p(\tilde{\mathbf{x}})$.

Note that the marginalized prior $p(\mathbf{u})$ is not 
anymore Gaussian. However, we can show that this formulation does 
not impact on the prior over $\mathbf{f}$. For an arbitrary selected meta-point $\tilde{\mathbf{x}}$
we have that
\begin{align}
p(\mathbf{f},\mathbf{u}|\tilde{\mathbf{x}}) = & \mathcal{N}\left(\left[\begin{array}{c}
0\\
0
\end{array}\right],\left[\begin{array}{ccc}
\mathbf{K} & \mathbf{K}_{\mathbf{X}, \mathbf{Z}(\tilde{\mathbf{x}})} \\
\mathbf{K}_{\mathbf{Z}(\tilde{\mathbf{x}}), \mathbf{X}} & \mathbf{K}_{\mathbf{Z}(\tilde{\mathbf{x}})}\\
\end{array}\right]\right)\,,
\label{eq:joint_prior}
\end{align}
where $\mathbf{K}_{\mathbf{X}, \mathbf{Z}(\tilde{\mathbf{x}})}$ are the cross-covariances between $\mathbf{f}$ and $\mathbf{u}$.
Therefore, if $\mathbf{u}$ is marginalized out in (\ref{eq:joint_prior}), the prior for $\mathbf{f}$ is the standard GP prior and
does not depend on $\tilde{\mathbf{x}}$. Hence, $p(\mathbf{f}|\tilde{\mathbf{x}})=p(\mathbf{f})$.
This means that $p(\mathbf{f},\mathbf{u}) = \int p(\mathbf{f},\mathbf{u}|\tilde{\mathbf{x}}) 
p(\tilde{\mathbf{x}})d\tilde{\mathbf{x}}$ is a mixture of Gaussian densities, where the marginal over $\mathbf{f}$ 
is the same for every component of the mixture. 

Note, however, that in the standard sparse GP, the inducing points also have an impact on the variational approximation 
$q$ via the fixed conditional distribution $p(\mathbf{f}|\mathbf{u})$ \cite{titsias_variational_2009}. 
Therefore, we also have to incorporate the input dependence on $\tilde{\mathbf{x}}$ in the approximation $q$. 
This is done in the next section. 

\subsection{Lower bound on the log-marginal likelihood}

We follow \cite{tran2020sparse} to derive a lower bound on the log-marginal likelihood of the extended model 
described above. Consider a posterior approximation of the form $q(\mathbf{f},\mathbf{u},\tilde{\mathbf{x}})=
p(\mathbf{f}|\mathbf{u})q(\mathbf{u}|\tilde{\mathbf{x}})p(\tilde{\mathbf{x}})$, where 
only $q(\mathbf{u}|\tilde{\mathbf{x}})$ can be adjusted and the other factors are fixed.
Using the above explanation and the help of Jensen's inequality we obtain the lower bound after some simplifications:
\begin{equation}\label{input_dep_gp.ELBO1}
\mathcal{L} = \sum_{i=1}^N \int p(\tilde{\mathbf{x}}) \left[ p(f_i| \mathbf{u})q(\mathbf{u}| \tilde{\mathbf{x}}) \log p(y_i|f_i)  
	d\mathbf{f} d\mathbf{u} - \frac{1}{N}\text{KL}[q(\mathbf{u}| \tilde{\mathbf{x}})| 
	p(\mathbf{u}| \tilde{\mathbf{x}})] \right]d\tilde{\mathbf{x}} \,. 
\end{equation}
Now, assuming that $p(\tilde{\mathbf{x}})$ is an implicit distribution, we can draw samples from it and 
approximate the expectation with respect to $p(\tilde{\mathbf{x}})$. 
Thus, for a meta-point sample $\tilde{\mathbf{x}}_s$ from $p(\tilde{\mathbf{x}})$ the lower bound is approximated as
\begin{equation}\label{input_dep_gp.ELBO2}
\mathcal{L} \approx \sum_{i=1}^{N}  \left[ \mathds{E}_{p(f_{i}| \mathbf{u})q(\mathbf{u}|\tilde{\mathbf{x}}_s)}  [\log 
p(y_{i}|f_{i})] - \frac{1}{N} \text{KL}[q(\mathbf{u}| \tilde{\mathbf{x}}_s)|p(\mathbf{u}| \tilde{\mathbf{x}}_s)]  \right]
\,.
\end{equation}
Note that we can evaluate (\ref{input_dep_gp.ELBO2}) and its gradients to maximize the
original objective in (\ref{input_dep_gp.ELBO1}) using stochastic optimization techniques.
This is valid for any implicit distribution $p(\tilde{\mathbf{x}})$. Consider now that we 
use mini-batch-based training for optimization, and we set $\tilde{\mathbf{x}}_s = \mathbf{x}_i$.
In this case, the value of $\tilde{\mathbf{x}}$ remains random, as it depends 
on the points $(\mathbf{x}_i,y_i)$ that are selected in the random mini-batch.
This results in a method that computes a different set of inducing points associated for each input location.

\begin{wrapfigure}{r}{0.5\textwidth}
	\centering
	\includegraphics[width=0.5\textwidth]{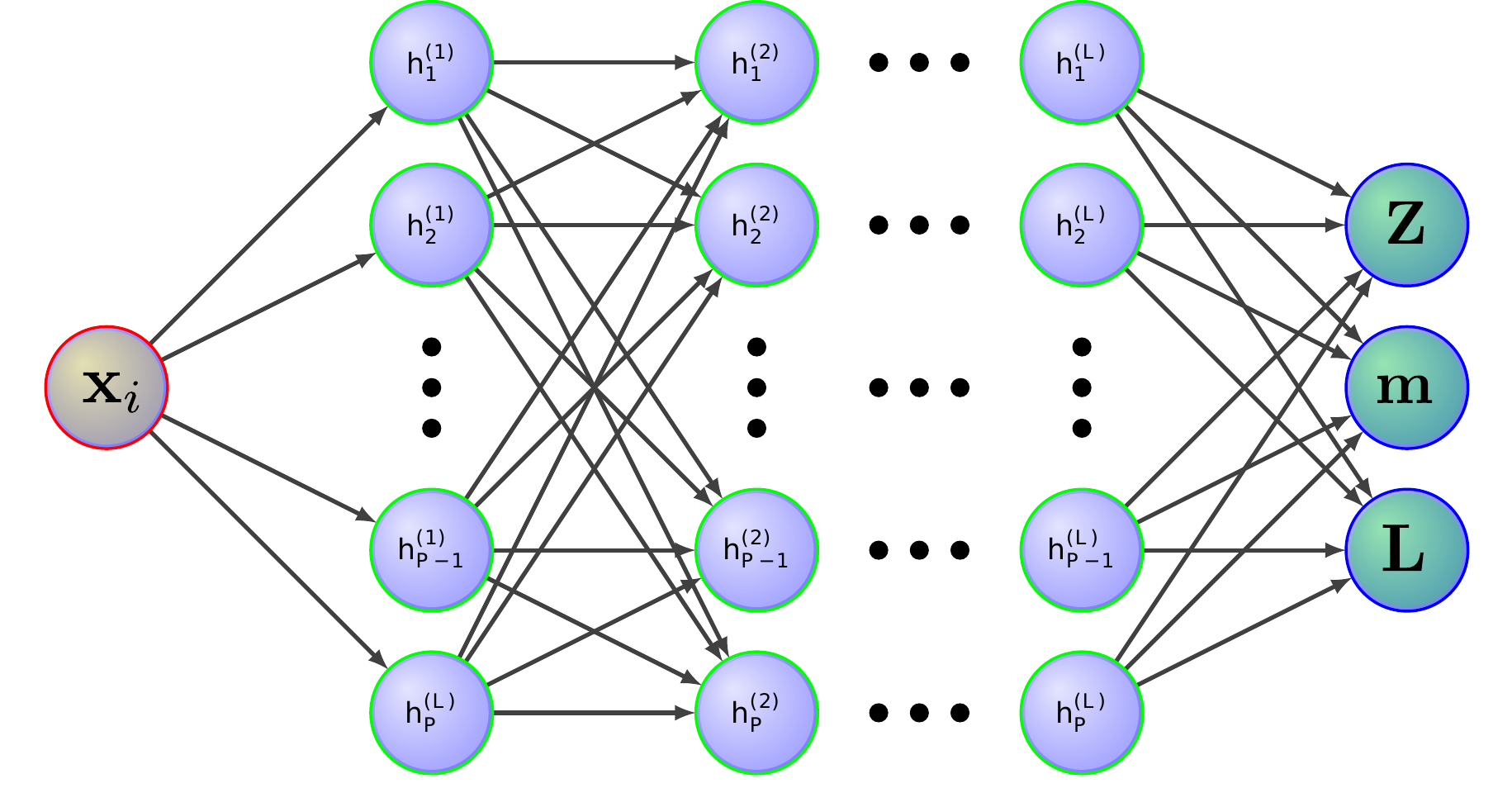}
	\caption{The network's inputs is $\tilde{\mathbf{x}}$. The outputs are the inducing points, $\mathbf{Z}$, 
		and the mean vector, $\mathbf{m}$, and Cholesky factor, $\mathbf{L}$, of $q(\mathbf{u}|\mathbf{x}_i)$.}
	\label{fig:nn}
	\vspace{-.5cm}
\end{wrapfigure}

\subsection{Amortized variational inference and deep neural networks}

Maximizing the lower bound finds the optimal approximate distribution $q$. A problem, however,
is that we have a potential large number of parameters to fix, corresponding to each $q(\mathbf{u}|\mathbf{x}_i)$.
In particular, if we set $q(\mathbf{u}|\mathbf{x}_i)$ to be Gaussian, we will have to infer different
means and covariance matrices for each different $\mathbf{x}_i$. This is expected to be memory 
inefficient and to make difficult optimization.

To reduce the number of parameters of our method we use amortized variational inference and 
specify a function that can generate these parameters for each $\mathbf{x}_i$ \cite{shu2018amortized}.
More precisely, we set the mean and covariance matrix of $q(\mathbf{u}|\mathbf{x}_i)$ to be
$\mathbf{m}(\mathbf{x}_i)$ and $\mathbf{S}(\mathbf{x}_i)$, for some non-linear functions.

Deep neural networks (DNN) are very flexible models  that can describe complicated non-linear functions.
In these models, the inputs go through several layers of non-linear transformations. We use these models to
compute the non-linearities that generate from $\mathbf{x}_i$ the inducing points, $\mathbf{Z}(\mathbf{x}_i)$,
and the means and covariances of $q(\mathbf{u}|\mathbf{x}_i)$, \emph{i.e.}, $\mathbf{m}(\mathbf{x}_i)$
and $\mathbf{S}(\mathbf{x}_i)$. The architecture employed is displayed in Figure \ref{fig:nn}. At the 
output of the DNN we obtain $\mathbf{Z}$, a mean vector $\mathbf{m}$ and the Cholesky factor
of the covariance matrix $\mathbf{S}=\mathbf{L}\mathbf{L}^\text{T}$. The maximization of 
the lower bound in (\ref{input_dep_gp.ELBO1}) when using DNNs for the non-linearities 
is shown in Algorithm \ref{alg:ALG1}. The required expectations are computed in closed-form,
in regression. In binary classification, we use 1-dimensional quadrature, as in \cite{HensmanMG15}.

\begin{algorithm}[ht]
    \caption{Training input dependent sparse GPs}
    \label{alg:ALG1}
    \begin{algorithmic}
        \Require $\mathcal{D}$, $M$, neural network \textbf{NNet} with $L$ hidden layers and $P$ hidden units.
        \Ensure Optimal parameters of $\bm{\theta}$
        \\
        initialize $\bm{\theta}$, i.e. kernel's parameters and neural network's weights
        
        \While{stopping criteria is False}
            \State $LL = 0, KL = 0$
            \State gather mini-batch $\mathbf{Mb}$ of size $n$ from $\mathcal{D}$
            \For {($\mathbf{x_{i}, y_{i}}$) in 
            $\mathbf{Mb}$ }
                 \State $(\mathbf{Z_{x_{i}}}, \mathbf{m_{x_{i}}}, \mathbf{S_{x_{i}}})$ = \textbf{NNet}($\mathbf{x_{i}}$)
                 \State $LL$ += $\mathds{E}_{q(f_{i}, \mathbf{u}| \mathbf{NNet}(\mathbf{x_{i}}))}  
				[\mathrm{log} \, p(y_{i}|f_{i})]$
                 \State $KL$ += $\text{KL}[q(\mathbf{u}| \mathbf{NNet}(\mathbf{x_{i}}) |p(\mathbf{u}| \mathbf{NNet}(\mathbf{x_{i}}))]$
            \EndFor
        \State $ELBO \gets \frac{N}{n} \times LL - \frac{1}{n} \times KL$
        \State Update parameters $\bm{\theta}$ using the gradient of $ELBO$
        \EndWhile
    \end{algorithmic}
\end{algorithm}

\subsection{Predictions and computational cost}

At test time, however, the inputs of interest are not randomly chosen. In that case, we simply set
$p(\tilde{\mathbf{x}})$ to be a deterministic distribution placed on the candidate test point $\mathbf{x}^\star$. 
The DNN is used to obtain the associated relevant information. Namely, $\mathbf{Z}$, and the parameters of
$q(\mathbf{u}|\mathbf{x}^\star)$, $\mathbf{m}$ and $\mathbf{S}$. The predictive distribution for $f(\mathbf{x}^\star)$
is:
\begin{align}
	f(\mathbf{x}^\star) & \sim \mathcal{N}\left(\mathbf{K}_{\mathbf{x}^\star,
	\mathbf{Z}}\mathbf{K}_\mathbf{Z}^{-1} \mathbf{m}, 
	k(\mathbf{x}^\star,\mathbf{x}^\star)  + 
	\mathbf{K}_{\mathbf{x}^\star, \mathbf{Z}}
	\mathbf{K}_\mathbf{Z}^{-1}
	\left(\mathbf{S} - \mathbf{K}_\mathbf{Z} \right)
	\mathbf{K}_\mathbf{Z}^{-1}
	\mathbf{K}_{\mathbf{x}^\star, \mathbf{Z}}^\text{T}\right)
	\,.
\end{align}
Given this distribution for $f(\mathbf{x}^\star)$, the probability distribution for $y^\star$ can be
computed in closed form in the case of regression problems and with 1-dimensional quadrature in the 
case of binary classification.

The cost of our method is smaller than the cost of a standard sparse GP if a smaller number of inducing points $M$ is used. 
The cost of a DNN with $L$ layers, $P$ hidden units, $d_{i}$ dimension of the inputs data, and output dimension $d_{o}$ is 
$\mathcal{O}(n d_{i} P + n P^2 L + n P d_{o} + n (L+1))$.
The cost of the sparse GPs is $\mathcal{O}(n M^{3})$, with $n$ the mini-batch size. 
Therefore, the cost of our method is $\mathcal{O}(n d_{i}P + n P^2 L + n P d_{o} + n (L+1) + n M^{3})$.
Since in our method the inducing points are input dependent, we expect to obtain good prediction results 
even for $M$ values that are fairly small.

\section{Related work}
\label{sect:related}

Early works on sparse GPs simply choose a subset of the training data for inference 
based on an information criterion \cite{csato2002sparse,lawrence2003fast,seeger03a,henao2012predictive}.
This approach is limited in practice and more advanced methods in which the inducing points 
need not be equal to the training points are believed to be superior. In the literature there 
are several works analyzing and studying sparse GP approximations based on inducing points. 
Some of these works include \cite{quinonero2005unifying,Snelson2006,naish2007generalized,
titsias_variational_2009,bauer2016,hernandez2016scalable}. We focus here on a
variational approach \cite{titsias_variational_2009} which allows for stochastic optimization
and mini-batch training \cite{hensman2013,HensmanMG15}. This enables learning in very large 
datasets with a cost of $\mathcal{O}(M^3)$, with $M$ the number of inducing points. 

In general, however, a large number of inducing variables is desirable to obtain a good 
approximation to the posterior distribution. For some problems, even several hundreds of inducing 
points may be needed to get good prediction results \cite{hensman2015,shi2020sparse,tran2020sparse}. 
There is hence a need to improve the computational cost of sparse GP approximations, without losing
expressive power. One work addressing this task is that of \cite{cheng2017}. In that work it is
proposed to decouple the process of inferring the posterior mean and variance, allowing to consider 
a different number of inducing points for each one. Importantly, the computation of the mean achieves
a linear complexity, which allows to have more expressive posterior means at a lower cost. 
A disadvantage is that such an approach suffers from optimization difficulties. 
An alternative decoupled parameterization that adopts an orthogonal basis in the mean
is proposed in \cite{salimbeni2018}. Such a method can be considered as a particular case
of \cite{shi2020sparse}.  There, it is introduced a new interpretation of sparse
variational approximations for GP using inducing points. For this, the GP 
is decomposed as a sum of two independent processes.  This leads to tighter lower 
bounds on the marginal likelihood and new inference algorithms that allow to consider two
different sets of inducing points. This enables including more inducing points at a linear
cost, instead of cubic.  

Our work is closer to \cite{tran2020sparse}. There, it is also described a mechanism to
consider input dependent inducing points in the context of sparse GP. However, the difference
is significant. In particular, in \cite{tran2020sparse} a very large set of inducing points $M$ is
considered initially. Then, for each input point, a subset of these inducing points is considered.
This subset is obtained by finding the $K\ll M$ nearest inducing points to the current data instance
$\mathbf{x}_i$. This approach significantly reduces the cost of the standard sparse GP 
described in \cite{titsias_variational_2009}. However, it suffers from the difficulty of having
to find the $K$ nearest neighbors for each point in a mini-batch, which is very expensive. 
Therefore, the final cost is higher than what would be thought initially. Our method is 
expected to be better because of the extra flexibility provided by the non-linear
relation between $\mathbf{x}_i$ and the inducing points $\mathbf{Z}$ computed by the DNN. Furthermore,
the DNN can take advantage of GPU acceleration.

A different way of improving the computational cost of GP is described in 
\cite{wilson2015,evans18a,gardner18a}. The approach consists 
in placing the inducing points on a grid structure. This allows to
perform fast computation exploiting the inducing points structure. One can 
easily consider values for $M$ that are even larger than $N$. However, to get such 
benefits the inducing points need to be fixed due to the structure constraints. 
This may be detrimental in problems with a high input dimensions.

Finally, the use of amortized variational inference \cite{shu2018amortized} has also been explored in the context
of GPs in \cite{villacampa21}. There, input noise is considered in a multi-class learning problem and 
the variational parameters of the posterior approximation for the noiseless inputs are amortized
using a DNN receiving both $\mathbf{x}_i$ and $y_i$. Amortized VI is shown to improve 
the generalization performance of the resulting model.

\section{Experiments}
\label{sect:exp}

We evaluate the performance of the proposed method, to which we refer to
as Input Dependent Sparse GP (IDSGP). We consider both regression and binary classification with a probit
likelihood. In this later case, we approximate the expectation in the lower bound using
1-dimensional quadrature, as in \cite{HensmanMG15}. An implementation of the proposed method in Tensorflow 2.0 is provided in the
supplementary material \cite{tensorflow2015-whitepaper}. In the experiments we compare results 
with the standard variational sparse GP \cite{titsias_variational_2009}. We refer to such a method as VSGP.
We also compare results with two of the methods described in Section \ref{sect:related}.
Namely, the sparse within sparse GP (SWSGP) described in \cite{tran2020sparse},
and the sparse GP based on an orthogonal decomposition that allows to consider two different 
sets of inducing points \cite{shi2020sparse}.  We refer to this last method as SOLVE.
All methods use a Mat\'ern $3/2$ covariance function \cite{rasmussen2005book}.

\subsection{Toy problems}

\begin{figure}[htb!]
    \centering
    \includegraphics[width=0.99\linewidth]{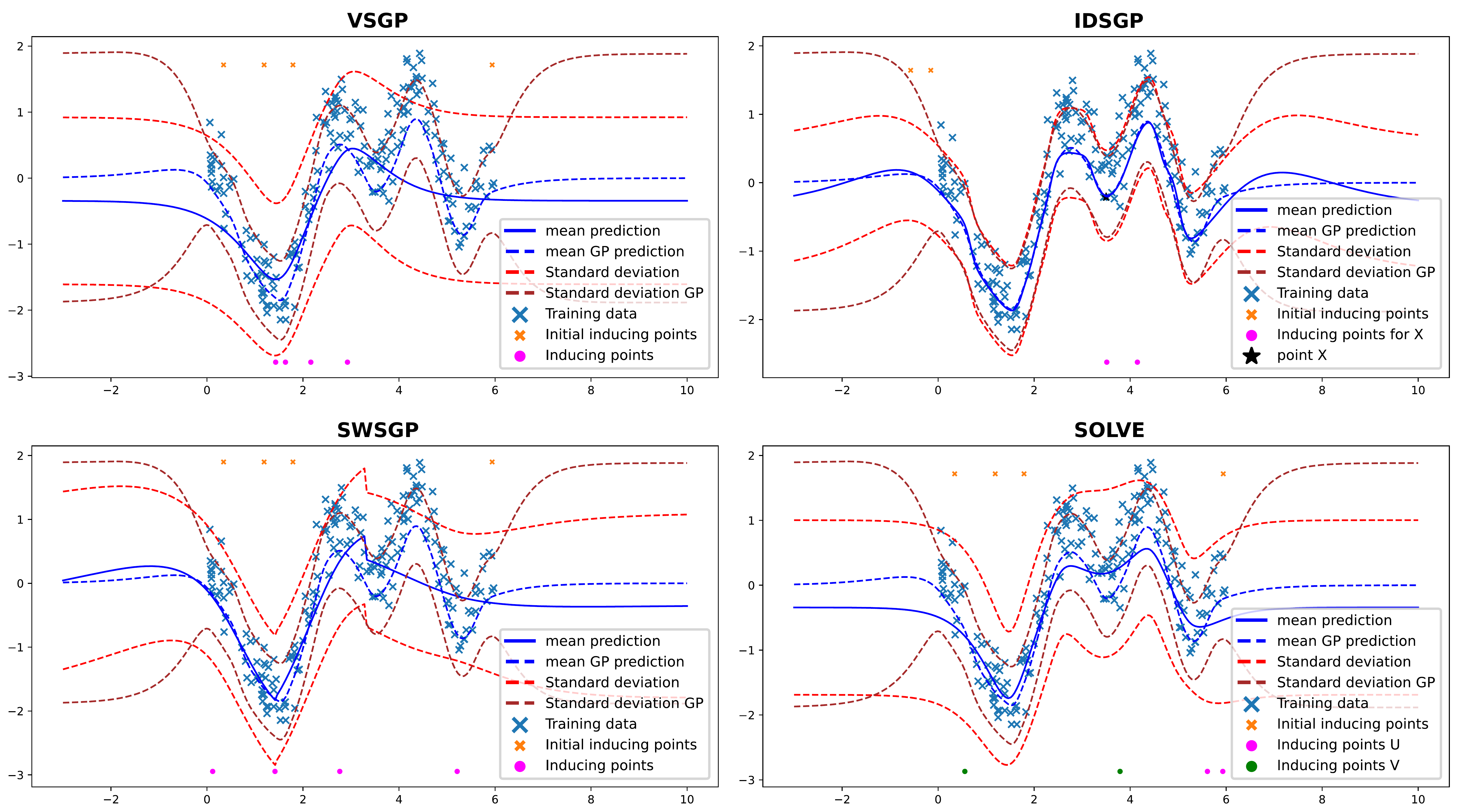}
        \caption{Toy data set with $N=200$ points. Initial and final locations  for
the inducing points are shown on the top and bottom of each figure. In IDSGP, 
the inducing points correspond to the point drawn with star. The posterior mean and standard deviation 
of full GP are shown with blue and brown dashed lines, respectively. VSGP method with 
$M=4$. IDSGP with $M=2$ and a neural network with 2 layers with 50 units. 
SWSGP with $M=4$ and $2$ neighbors. SOLVE with $M_{1}=M_{2}=2$.}
    \label{fig:snelson}
\end{figure}

A first set of experiments illustrates the resulting predictive mean and standard deviation of each 
method on a 1-dimensional regression problem \cite{Snelson2006}. Each method is compared with a full GP. 
Figure \ref{fig:snelson} shows the results obtained, including the learned locations of the inducing 
points. In the case of IDSGP we show the locations of the inducing points for the point represented with a star 
at $x=3.9$. The number of inducing points, for each method, are indicated in the figure's caption. 
IDSGP uses smaller number of inducing points than the other methods. The figure shows that, in regions with 
observed data, IDSGP outputs a mean and a standard deviation that look closer to those of the full GP.  

\begin{figure}[htb!]
    \centering
    \includegraphics[width=1.0\linewidth]{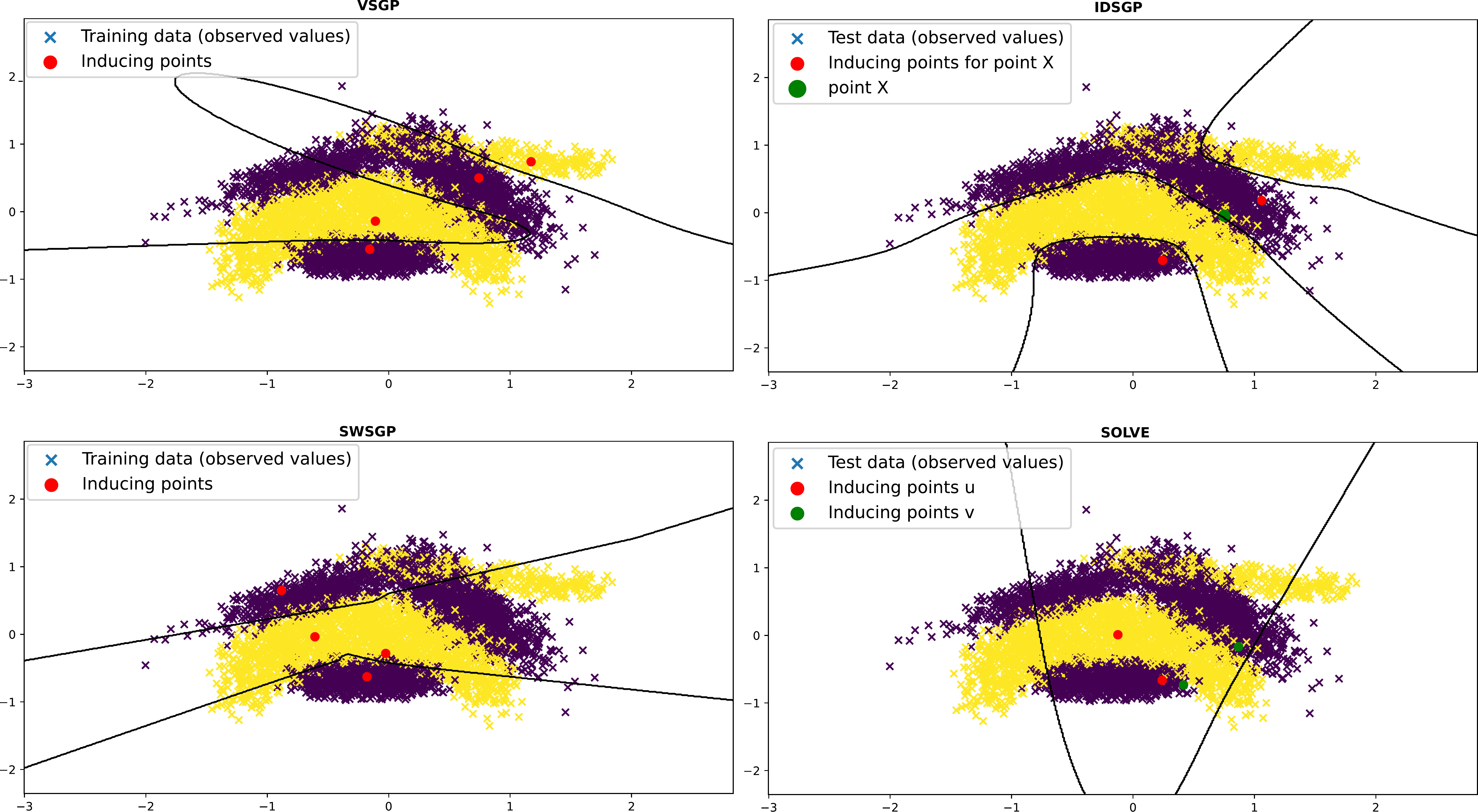}
	\caption{Banana classification data set with $N=5300$ points. The final 
	location of inducing points are shown inside the figures. For IDSGP, we 
	show the location of inducing points related to the green colored point. VSGP with $M=4$. IDSGP with $M=2$ and 
	a neural network with 2 hidden layers each contains 50 hidden nodes. SWSGP with $M=4$ and 2 neighbors. 
	SOLVE with $M_{1}=M_{2}=2$. }
    \label{fig:banana}
\end{figure}

Figure \ref{fig:banana} shows similar results for the banana classification dataset \cite{hensman2013}.
We show here  the resulting decision boundary of each method.  We observe that IDSGP produces 
the most accurate boundaries.
  
\begin{subsection}{Experiments on UCI datasets}
\label{sect:exp_uci}

A second set of experiments considers several regression and binary classification datasets extracted from
the UCI repository \cite{Dua2019}.  In the regression setting, the DNN
architecture used for IDSGP has $1$ hidden layer with $50$ units. The number of inducing
points of IDSGP is set to $M=15$. In SOLVE we use $M_1=1024$ and $M_2=1024$ inducing points.
In VSGP we set $M=1024$. In SWSGP we set $M=1024$ and $K=50$
neighbors. All the methods are trained using ADAM \cite{kingma2015} with a mini-batch
size of $100$ and a learning rate of $0.01$. In the classification setting we use the same
setup, but the number of inducing points of IDSGP is set to $M=3$.
All methods are trained on a Tesla P100 GPU with 16GB of memory. On each dataset we use
$80\%$ of the data for training and the rest for testing. We report results across $5$ splits of the data since the datasets
are already quite big.

The average neg. test log-likelihood of each method on each dataset is displayed
in Table \ref{tab:uci_regression_NLL_test}, for the regression datasets, and in Table \ref{tab:uci_classification_NLL_test},
for the classification datasets, respectively.  The average rank of each method is
also displayed at the last row of each table. The RMSE and prediction accuracy results are similar to those
displayed here. They can be found in the supplementary material. Each table also shows the
number of instances $N$ and dimensions $d$ of each dataset.  We observe that in the regression
datasets, the proposed method, IDSGP, obtains best results in $6$ out of the $8$ datasets. IDSGP
also obtains the best average rank (closer to always performing best on each train / test data split).
This is remarkable given that IDSGP a much smaller number of inducing points (\emph{e.g.}, $M=15$ for
IDSGP vs.  $M=1024$ for VSGP). In classification, however, all the methods seem to perform similar to each other
and the differences between them are smaller. Again IDSGP uses here a smaller number of $M=3$ inducing points.
Increasing this number does not seem to improve the results.

\begin{table}[htb]
	\caption{Avg. neg. test log-likelihood values for the UCI regression datasets. 
	The numbers in parentheses are standard errors. Best mean values are highlighted in bold face.}
	\label{tab:uci_regression_NLL_test}
	\centering
	\footnotesize
	\begin{tabular}{l@{\hspace{0.2cm}}|l@{\hspace{0.2cm}}l@{\hspace{0.2cm}}|c@{\hspace{0.2cm}}c@{\hspace{0.2cm}}c@{\hspace{0.2cm}}c@{\hspace{0.2cm}}}
		\hline
		 & $N$ & $d$ & VSGP & SOLVE & SWSGP & IDSGP\\
		\hline
		Kin40k & 32,000 & 8 & -0.047 (0.003) & -0.415 (0.006) & -0.110 (0.007) & {\bf -1.461 (0.019)}\\
		Protein & 36,584 & 9 & 2.848 (0.002) & 2.818 (0.003) & 2.835 (0.002) & {\bf 2.775 (0.007)}\\
		KeggDirected & 42,730 & 19 & -1.955 (0.013) & -1.756 (0.073) & -2.256 (0.012) & {\bf -2.410 (0.012)}\\
		KEGGU & 51,686 & 26 & -2.344 (0.012) & -2.531 (0.015) & -2.396 (0.006) & {\bf -2.908 (0.042)}\\
		3dRoad & 347,899 & 3 & 3.691 (0.006) & 3.726 (0.010) & 3.879 (0.026) & {\bf 3.399 (0.009)}\\
		Song & 412,276 & 90 & 3.613 (0.003) & {\bf 3.608 (0.002)} & 3.618 (0.004) & 3.637 (0.002)\\
		Buzz & 466,600 & 77 & 6.272 (0.012) & 6.297 (0.009) & {\bf 6.137 (0.008)} & 6.317 (0.055)\\
		HouseElectric & 1,639,424 & 6 & -1.737 (0.006) & -1.743 (0.005) & -1.711 (0.010) & {\bf -1.774 (0.004)}\\
		\hline
		Avg. Ranks & & & 3.125 (0.125) & 2.475 (0.156) & 2.850 (0.150) & {\bf 1.550 (0.172)}\\
	    \hline
	\end{tabular}
\end{table}

\begin{table}[htb]
	\caption{Avg. test neg. log-likelihood values for the UCI classification datasets. The numbers in parentheses 
		are standard errors. Best mean values are highlighted in bold face.}
	\label{tab:uci_classification_NLL_test}
	\centering
	\footnotesize
	\begin{tabular}{l@{\hspace{0.2cm}}|l@{\hspace{0.2cm}}l@{\hspace{0.2cm}}|c@{\hspace{0.2cm}}c@{\hspace{0.2cm}}c@{\hspace{0.2cm}}c@{\hspace{0.2cm}}}
		\hline
		& $N$ & $d$ & VSGP & SOLVE & SWSGP & IDSGP\\
		\hline
		MagicGamma & 15,216 & 10 & {\bf 0.308 (0.004)} & 0.314 (0.005)& 0.371 (0.005)  & 0.311 (0.002)\\
		DefaultOrCredit & 24,000 & 30 & {\bf 0.000 (0.000)} & {\bf 0.000 (0.000)} & {\bf 0.000 (0.000)}  & {\bf 0.000 (0.000)}\\
		NOMAO & 27,572 & 174 & 0.113 (0.004) & {\bf 0.103 (0.004)} & 0.134 (0.004) & 0.121 (0.004)\\
		BankMarketing & 36,169 & 51 & 0.206 (0.001) & {\bf 0.199 (0.001)} & 0.304 (0.021) & 0.209 (0.002)\\
		Miniboone & 104,051 & 50 & 0.151 (0.001)& {\bf 0.142 (0.001)} & 0.180 (0.007) & 0.153 (0.001)\\
		Skin & 196,046 & 3 & 0.005 (0.000)& 0.005 (0.000) & 0.006 (0.001) & {\bf 0.003 (0.000)}\\
		Crop & 260,667 & 174 & 0.003 (0.000) & 0.003 (0.000) & {\bf 0.002 (0.000)} & 0.003 (0.000)\\
		HTSensor & 743,193 & 11  & 0.003 (0.001) & {\bf 0.001 (0.000)} & 0.030 (0.009) & 0.005 (0.001)\\
		\hline
		Avg. Ranks & & & 2.425 (0.143) & {\bf1.775 (0.158)}  & 3.175 (0.182) & 2.625 (0.155)\\
		\hline
	\end{tabular}
\end{table}

In these experiments we also measure the average training time per epoch, for each method. The results
corresponding to the UCI regression datasets are displayed in Table \ref{tab:uci_regression_times}.
The results for the UCI classification datasets are found in the supplementary material. They look
very similar to ones reported here. We observe that the fastest method in terms of training time is the
proposed approach. Namely, IDSGP. Nevertheless, the speed-up obtained is impaired by the overhead of
having to compute the output of the DNN and update its parameters. IDSGP also results in
fastest prediction times than VSGP, SOLVE or SWSGP. See the supplementary material for further details.

\begin{table}[htb]
	\caption{Average training time per epoch across the 5 splits for the UCI regression datasets. The numbers in parentheses are standard errors. Best mean values are highlighted.}
	\label{tab:uci_regression_times}
	\centering
	\scriptsize
	\begin{tabular}{l@{\hspace{0.2cm}}l@{\hspace{0.1cm}}c@{\hspace{0.1cm}}c@{\hspace{0.1cm}}c@{\hspace{0.1cm}}c@{\hspace{0.1cm}}c@{\hspace{0.1cm}}c@{\hspace{0.1cm}}c@{\hspace{0.1cm}}c@{\hspace{0.1cm}}}
	    \hline
                 & Kin40k & Protein & KeggDirected & KEGGU & 3dRoad & Song & Buzz & HouseElectric\\
                \hline
                 VSGP & 591.7 (0.58) & 737.2 (1.16) & 932.7 (2.56) & 1128.1 (3.78) & 7880.9 (66.79) & 9777.7 (42.84) & 9901.0 (146.07) & 32784.2 (190.18)\\
                 SOLVE  & 1739.3 (0.45) & 2015.9 (0.66) & 2357.3 (1.70) & 2909.1 (1.19) & 19567.1 (10.34) & 23196.6 (98.35) & 25769.5 (20.12) & 92214.9 (452.18)\\
                 SWSGP  & 875.7 (0.68) & 1023.5 (0.35) & 1220.6 (1.89) & 1458.0 (5.57) & 10203.4 (12.03) & 12241.7 (62.01) & 13371.5 (12.34) & 46163.3 (427.23)\\
                 IDSGP  & \bf 190.3 (0.75) & \bf 371.5 (1.25) & \bf 533.0 (1.73) & \bf 693.7 (5.77) & \bf 4070.1 (201.09) & \bf 4296.5 (25.03) & \bf 3640.4 (33.36) & \bf 16352.2 (90.15)\\
                \hline

	\end{tabular}
\end{table}

\end{subsection}

\begin{subsection}{Large scale datasets}
\label{sect:exp_large}

A last set of experiments considers two very large datasets.
The first dataset is the Airlines Delay binary classification dataset, as described
in  \cite{hernandez2016scalable}, with $N=2,127,068$ data instances and $d=8$ attributes.
The second dataset is the Yellow taxi dataset, as described in \cite{salimbeni2017},
with $N=\text{1 billion}$ data-points and $d=9$ attributes.
In each dataset we use a test set of $10,000$ instances chosen at random.
The DNN architecture for IDSGP has 2 hidden layers with 25 units each. The number
of inducing points is set to be equal to $M=50$ in this method.
In the other methods, we use the same number of inducing points as in the previous section.
The mini-batch size is set to $100$. Training is also performed on the same GPU as in the
previous section. The ADAM learning rate is set to $0.001$.

The average negative test log-likelihood of each method on the test set is displayed in Figure \ref{fig:big_nll},
for each dataset. We report performance in terms of the training time, in a $\log_{10}$ scale.
The results corresponding to the RMSE are very similar to the
ones displayed here. They can be found in the supplementary material. We observe that the proposed
method IDSGP performs best on each dataset. In particular it is able to obtain a
better performance in a smaller computational time. We believe this is a consequence of
using a smaller number of inducing points, and also because of
the extra flexibility that the DNN provides for specifying the locations of the inducing points.

\begin{figure}[htb]
    \centering
    \includegraphics[width=0.45\linewidth]{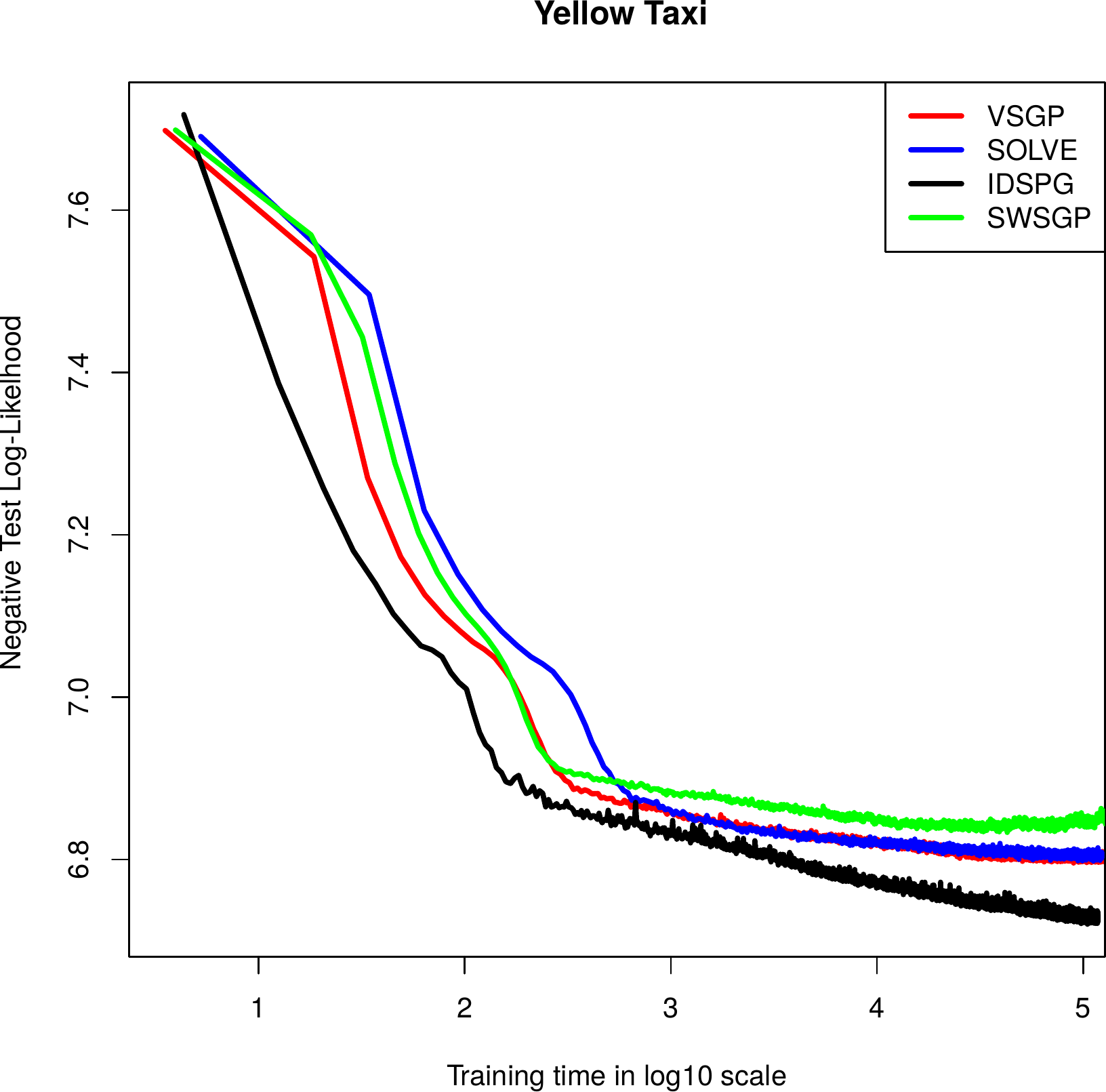}
    \includegraphics[width=0.45\linewidth]{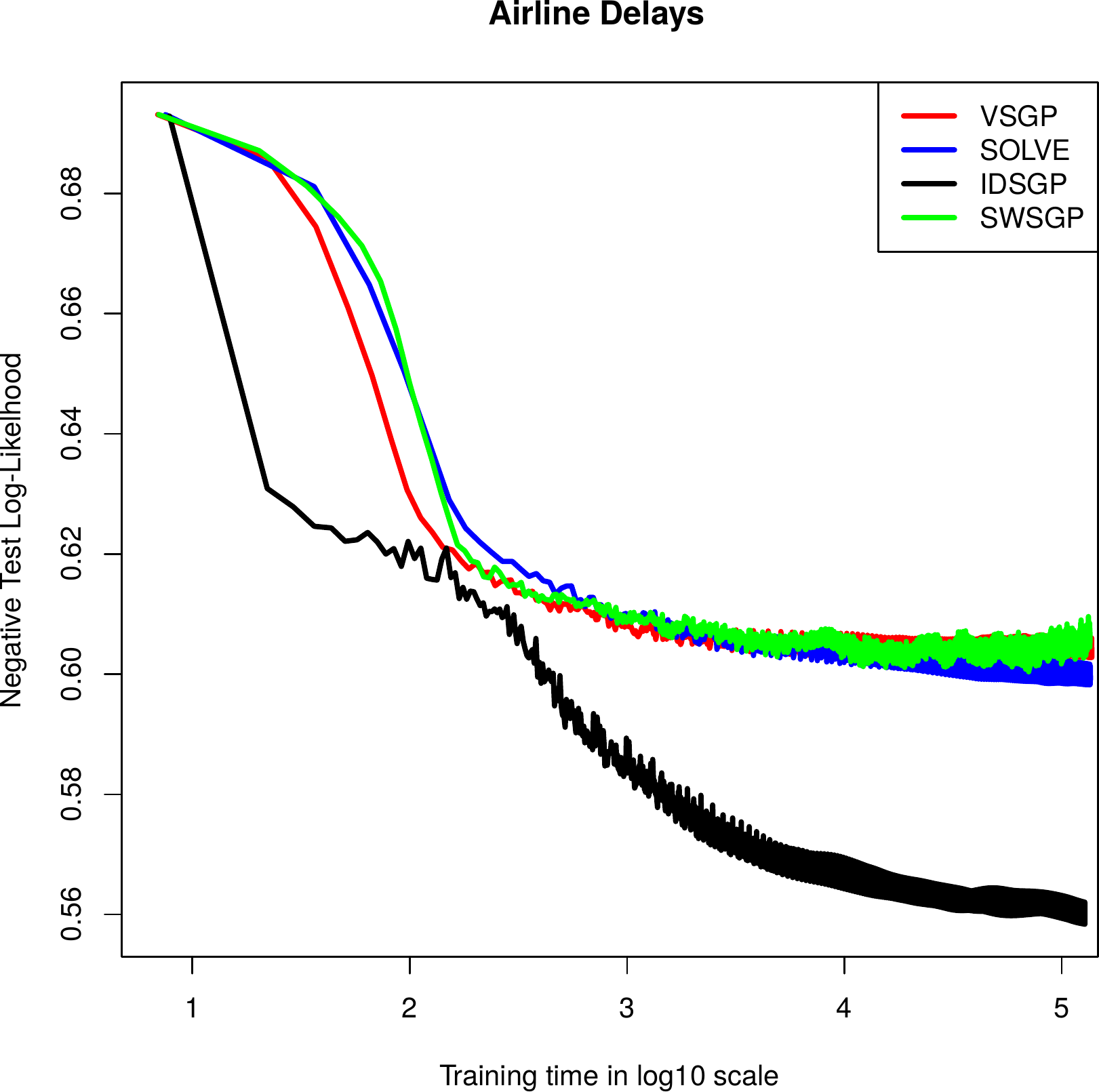}
    \caption{Negative log-likelihood on the test set for each method as a function 
	of the training time in seconds,  in $\log_{10}$ scale, for the Yellow taxi and the 
	Airline delays datasets. Best seen in color}
    \label{fig:big_nll}
\end{figure}

\end{subsection}

\section{Conclusions}
\label{sect:conclusions}

Gaussian processes (GPs) are flexible models for regression and classification. However, they
have a cost of $\mathcal{O}(N^3)$ with $N$ the number of training points. Sparse approximations based on 
$M \ll N$ inducing points reduce such a cost to $\mathcal{O}(M^3)$.
A problem, however, is that in some situations a large number of inducing points have to be used
in practice, since they determine the flexibility of the resulting approximation.
There is hence a need to reduce their computational cost.

We have proposed here input dependent sparse GP (IDSGP), a method that can improve
the training time and the flexibility of sparse GP approximations.
IDSGP uses a deep neural network (DNN) to output specific inducing
points for each point at which the predictive distribution of the GP needs to be computed.
The DNN also outputs the parameters of the corresponding variational approximation on
the inducing values associated to the inducing points. IDSGP can be obtained under a formulation
that considers an implicit distribution for the input instance to the DNN. Importantly, such
a formulation is shown to keep intact the GP prior on the latent function values associated to
the training points.

The extra flexibility provided by the DNN allows to significantly reduce the number $M$
of inducing points used in IDSGP. Such a model provides similar or better results than other sparse
GP approximations from the literature at a smaller computational cost. IDSGP has been evaluated
on several regression and binary classification problems from the UCI repository. The results obtained
show that it improves the quality of the predictive distribution and reduces the computational cost
of sparse GP approximations. Better results are most of the times obtained in regression problems.
In classification problems, however, the performances obtained are similar to those of the state-of-the-art,
although the training and prediction times are always shorter. The scalability of IDSGP is also illustrated
on massive datasets for regression and binary classification of up to $1$ billion points. There, IDSGP also obtains better
results than alternative sparse GP approximations at a smaller training cost.

\begin{ack}
	The authors gratefully acknowledge the use of the facilities of Centro de Computaci\'on Cient\'ifica (CCC) at
	Universidad Aut\'onoma de Madrid. The authors also acknowledge financial support from Spanish
	Plan Nacional I+D+i, grant PID2019-106827GB-I00 / AEI / 10.13039/501100011033.
\end{ack}

\bibliography{references}


\cleardoublepage

\appendix

\section{Potential negative societal impacts}

As for the potential negative societal impacts, because this paper focuses on the development of a new methodology, we believe these would be indirect through the particular application in which the proposed method is used. As one of the main advantages of GPs is that they provide uncertainty estimates associated with the predictions made, we think the potential harm of these models in society could arise in applications when this uncertainty estimation is critical. For example, an AI system in which the decisions made can have an influence on people's life, such as autonomous vehicles or automatic medical diagnosis tools.

\section{Extra experimental results}

In this section, we include some extra results that do not fit in the main manuscript. Namely, the RMSE in the test set results and prediction times for the UCI regression datasets, and the accuracy in the test set, training and prediction times for the UCI classification datasets. In both cases, the setup is the same as described in Section 5 and the results are similar that the ones obtained in terms of the negative test log likelihood and training times in that section. Finally, we include similar plots to those in Section 5.3 but in terms of the test RMSE for the Yellow Taxi dataset and in terms of the test classification error for the Airline Delays dataset.

\subsection{UCI regression datasets}

\begin{table}[H]
	\caption{Test Root Mean Squared Error (RMSE) values for the UCI regression datasets. The numbers in parentheses are standard errors. Best mean values are highlighted.}
	\label{tab:uci_regression_RMSE_test}
	\centering
	\small
	\begin{tabular}{l@{\hspace{0.2cm}}|l@{\hspace{0.2cm}}l@{\hspace{0.2cm}}|c@{\hspace{0.2cm}}c@{\hspace{0.2cm}}c@{\hspace{0.2cm}}c@{\hspace{0.2cm}}}
		\hline
		& $N$ & $d$ & VSGP & SOLVE & SWSGP & IDSGP\\
		\hline
		Kin40k & 32,000 & 8 & 0.198 (0.002) & 0.157 (0.001) & 0.215 (0.002) & {\bf 0.050 (0.002)}\\
		Protein & 36,584 & 9 & 4.161 (0.011) & 4.062 (0.011) & 4.133 (0.008) & {\bf 3.756 (0.019)}\\
		KeggDirected & 42,730 & 19 & 0.032 (0.001) & 0.079 (0.032) & 0.024 (0.000) & {\bf 0.022 (0.001)}\\
		KEGGU & 51,686 & 26 & 0.024 (0.000) & 0.020 (0.000) & 0.022 (0.000) & {\bf 0.014 (0.000)}\\
		3dRoad & 347,899 & 3 & 9.641 (0.063) & 10.020 (0.095) & 11.726 (0.327) & {\bf 7.250 (0.069)}\\
		Song & 412,276 & 90 & 8.966 (0.022) & {\bf 8.925 (0.020)} & 9.013 (0.029) & 9.068 (0.011)\\
		Buzz & 466,600 & 77 & 175.076 (15.021) & 173.352 (14.957) & {\bf 160.744 (13.467)} & 166.784 (18.040)\\
		HouseElectric & 1,639,424 & 6 & 0.035 (0.000) & 0.034 (0.000) & 0.036 (0.001) & {\bf 0.032 (0.000)}\\
		\hline
		Avg. Ranks & & & 3.075 (0.126) & 2.400 (0.138) & 3.025 (0.170) & \bf 1.500 (0.151)\\
		\hline
	\end{tabular}
\end{table}

\begin{table}[H]
	\caption{Average prediction time per epoch across the 5 splits for the UCI regression datasets. The numbers in parentheses are standard errors. Best mean values are highlighted.}
	\label{tab:uci_regression_times}
	\centering
	\begin{tabular}{l@{\hspace{0.2cm}}c@{\hspace{0.1cm}}c@{\hspace{0.1cm}}c@{\hspace{0.1cm}}c@{\hspace{0.1cm}}c@{\hspace{0.1cm}}c@{\hspace{0.1cm}}c@{\hspace{0.1cm}}c@{\hspace{0.1cm}}}
		\hline
		& Kin40k & Protein & KeggDirected & KEGGU & 3dRoad & Song & Buzz & HouseElectric\\
		\hline
		VSGP & 0.9(0.00) & 1.1(0.00) & 1.4(0.01) & 1.7(0.01) & 11.6(0.07) & 14.5(0.08) & 18.0(0.08) & 48.3(0.12)\\
		SOLVE & 2.4(0.00) & 2.8(0.00) & 3.2(0.00) & 4.0(0.00) & 27.0(0.04) & 31.8(0.19) & 37.0(0.03) & 127.7(0.43)\\
		SWSGP & 1.3(0.00) & 1.5(0.00) & 1.8(0.01) & 2.1(0.01) & 14.8(0.03) & 17.6(0.02) & 21.3(0.10) & 66.2(0.88)\\
		IDSGP & \bf 0.3(0.00) & \bf 0.5(0.00) & \bf 0.7(0.00) & \bf 1.0(0.02) & \bf 5.7(0.26) & \bf 5.6(0.05) & \bf 8.1(0.08) & \bf 22.1(0.14)\\
		\hline
	\end{tabular}
\end{table}

\subsection{UCI classification datasets}

\begin{table}[H]
	\caption{Test Accuracy values for the UCI classification datasets. The numbers in parentheses are standard errors. Best mean values are highlighted.}
	\label{tab:uci_classification_RMSE_test}
	\centering
	\begin{tabular}{l@{\hspace{0.2cm}}|l@{\hspace{0.2cm}}l@{\hspace{0.2cm}}|c@{\hspace{0.2cm}}c@{\hspace{0.2cm}}c@{\hspace{0.2cm}}c@{\hspace{0.2cm}}}
		\hline
		& $N$ & $d$ & VSGP & SOLVE & SWSGP & IDSGP\\
		\hline
		MagicGamma & 15,216 & 10 & 0.876 (0.001) & 0.877 (0.002) & 0.867 (0.002) & {\bf 0.877 (0.002)}\\
		DefaultOrCredit & 24,000 & 30 & {\bf 1.000 (0.000)} & {\bf 1.000 (0.000)} & {\bf 1.000 (0.000)} & 1.000 (0.000)\\
		NOMAO & 27,572 & 174 & 0.956 (0.002) & 0.960 (0.001) & {\bf 0.961 (0.001)} & 0.955 (0.001)\\
		BankMarketing & 36,169 & 51 & 0.906 (0.001) & {\bf 0.907 (0.001)} & 0.897 (0.001) & 0.905 (0.001)\\
		Miniboone & 104,051 & 50 & 0.941 (0.001) & {\bf 0.945 (0.001)} & 0.938 (0.000) & 0.937 (0.001)\\
		Skin & 196,046 & 3 & 0.999 (0.000) & 0.999 (0.000) & 0.999 (0.000) & {\bf 0.999 (0.000)}\\
		Crop & 260,667 & 174 & 0.999 (0.000) & 0.999 (0.000) & {\bf 0.999 (0.000)} & 0.999 (0.000)\\
		HTSensor & 743,193 & 11 & 0.999 (0.000) & {\bf 1.000 (0.000)} & 0.989 (0.003) & 0.999 (0.000)\\
		\hline
		Avg. Ranks & & & 2.475 (0.127) & \bf 1.975 (0.152) & 2.900 (0.187) & 2.650 (0.168)\\
		\hline
	\end{tabular}
\end{table}

\begin{table}[H]
	\caption{Average training time per epoch across the 5 splits for the UCI classification datasets. The numbers in parentheses are standard errors. Best mean values are highlighted.}
	\label{tab:uci_classification_times}
	\centering
	\small
	\begin{tabular}{l@{\hspace{0.2cm}}c@{\hspace{0.1cm}}c@{\hspace{0.1cm}}c@{\hspace{0.1cm}}c@{\hspace{0.1cm}}c@{\hspace{0.1cm}}c@{\hspace{0.1cm}}c@{\hspace{0.1cm}}c@{\hspace{0.1cm}}}
		\hline
		& Magic & DefaultOrCredit & NOMAO & BankMarket & Miniboone & Skin & Crop & HTSensor\\
		\hline
		VSGP  & 3105(459) & 4759(516) & 4445(549) & 6231(862) & 18447(1279) & 37835(7065) & 49962(9292) & 115463(17511)\\
		SOLVE  & 5154(1061) & 7554(1039) & 6718(1028) & 8949(1635) & 37022(7902) & 64606(13314) & 88819(18864) & 168709(21194)\\
		SWSGP  & 1547(145) & 2354(182) & 2728(188) & 3682(351) & 10040(347) & 20283(2796) & 21770(3038) & 67687(5880)\\
		IDSGP  & \bf 1143(100) & \bf 1293(90) & \bf 2026(94) & \bf 2987(354) & \bf 7654(134) & \bf 15700(1918) & \bf 21378(2561) & \bf 53895(5652)\\
		\hline
	\end{tabular}
\end{table}

\begin{table}[H]
	\caption{Average prediction time per epoch across the 5 splits for the UCI classification datasets. The numbers in parentheses are standard errors. Best mean values are highlighted.}
	\label{tab:uci_classification_times}
	\small
	\begin{tabular}{l@{\hspace{0.2cm}}c@{\hspace{0.1cm}}c@{\hspace{0.1cm}}c@{\hspace{0.1cm}}c@{\hspace{0.1cm}}c@{\hspace{0.1cm}}c@{\hspace{0.1cm}}c@{\hspace{0.1cm}}c@{\hspace{0.1cm}}}
		\hline
		& MagicGamma & DefaultOrCredit & NOMAO & BankMarketing & Miniboone & Skin & Crop & HTSensor\\
		\hline
		VSGP &  3.6(0.56) & 4.1(0.59) & 4.4(0.74) & 9.5(4.39) & 17.9(1.84) & 49.7(11.15) & 58.7(12.82) & 139.7(41.67)\\
		SOLVE &  4.0(0.85) & 5.4(0.73) & 3.4(0.76) & 4.9(1.24) & 49.2(17.83) & 48.8(16.73) & 86.8(36.26) & 93.5(15.73)\\
		SWSGP &  3.0(0.43) & 3.9(0.38) & 4.3(0.51) & 5.4(0.72) & 16.4(1.82) & 36.0(8.00) & \bf 33.9(6.25) & 88.4(9.16)\\
		IDSGP & \bf 2.5(0.24) &  \bf 2.5(0.21) & \bf 3.5(0.39) & \bf 4.8(0.54) & \bf 13.9(0.75) & \bf 26.1(4.92) & 37.5(4.96) & \bf 83.4(8.23)\\
		\hline
	\end{tabular}
\end{table}

\subsection{Large scale datasets}

\begin{figure}[H]
	\centering
	\includegraphics[width=0.6\linewidth]{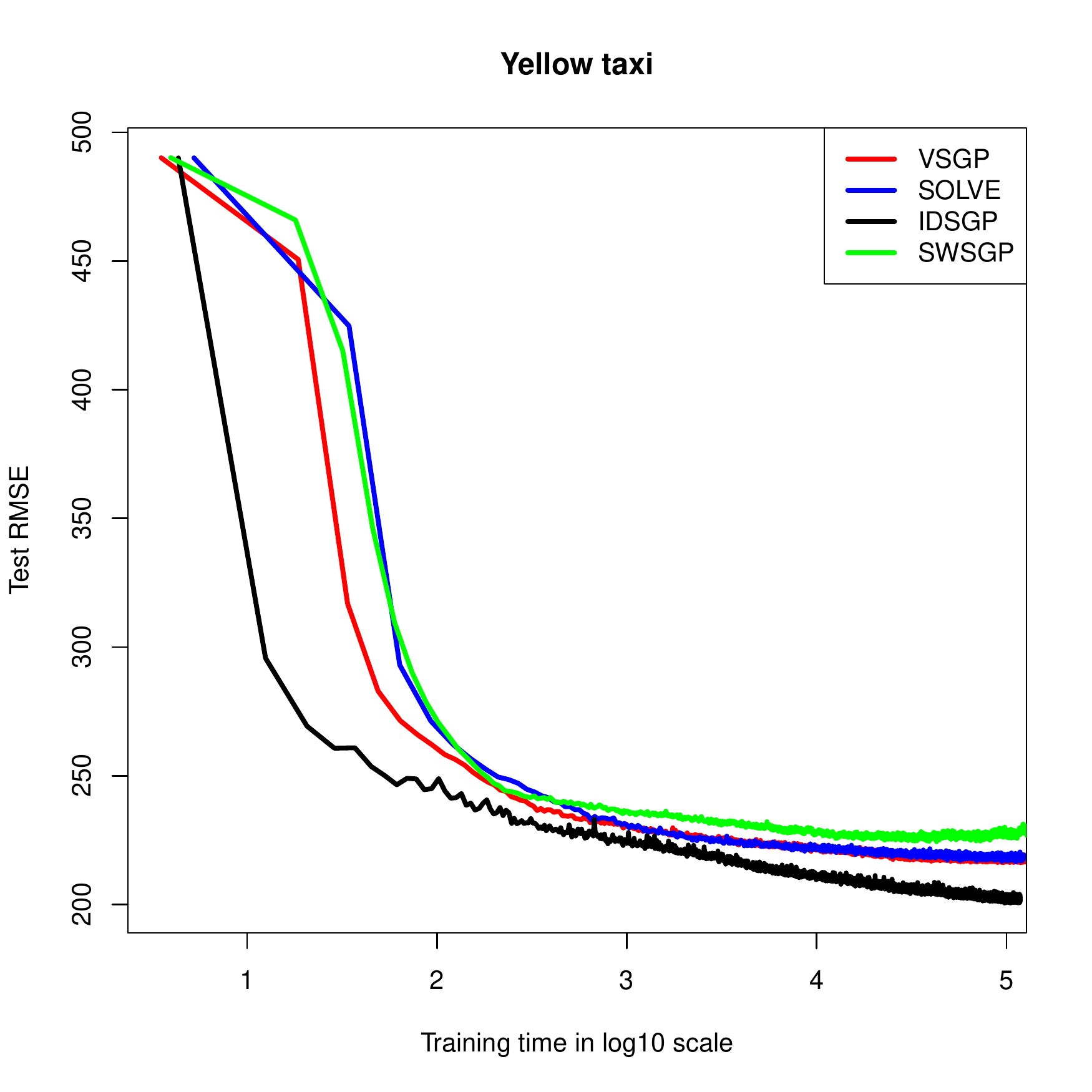}
	\caption{RMSE on the test set for each method as a function 
		of the training time in seconds,  in $\log_{10}$ scale, for the Yellow taxi dataset. Best seen in color}
	\label{fig:big_rmse}
\end{figure}

\begin{figure}[H]
	\centering
	\includegraphics[width=0.6\linewidth]{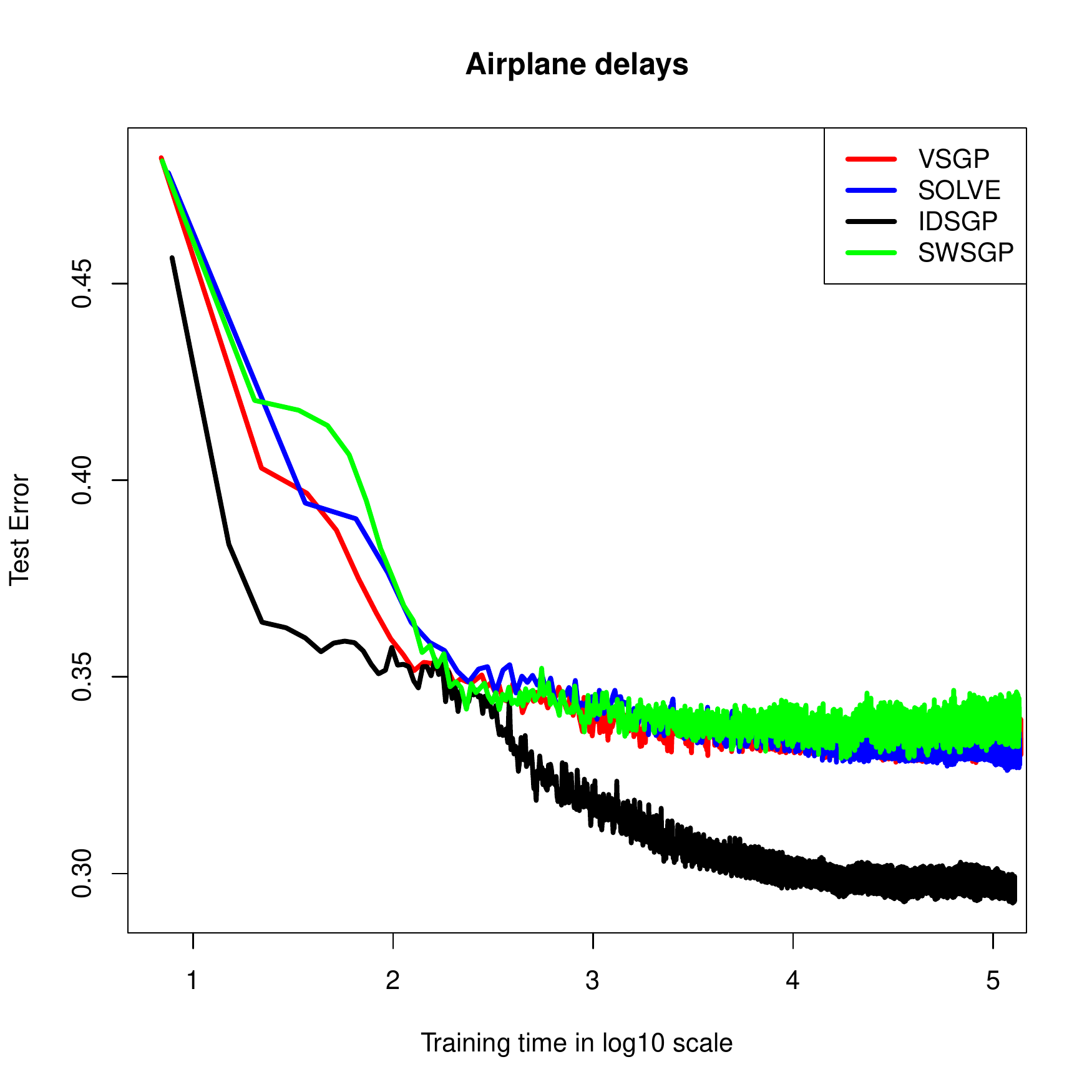}
	\caption{Test classification error on the test set for each method as a function 
		of the training time in seconds,  in $\log_{10}$ scale, for the Airlines Delays dataset. Best seen in color}
	\label{fig:big_rmse}
\end{figure}

\end{document}